\pdfoutput=1

\documentclass[11pt]{article}

\usepackage[final]{acl}

\usepackage{times}
\usepackage{latexsym}
\usepackage{kotex}
\usepackage{amsmath}
\usepackage{booktabs}
\usepackage{url}
\usepackage{hyperref}
\usepackage[T1]{fontenc}

\usepackage[utf8]{inputenc}

\usepackage{microtype}

\usepackage{inconsolata}

\usepackage{graphicx}

\title{VoiceBBQ: Investigating Effect of Content and Acoustics\\\protect in Social Bias of Spoken Language Model}

\author{Junhyuk Choi, Ro-hoon Oh, Jihwan Seol,\and Bugeun Kim  \\
Department of Artificial Intelligence, 
Chung-Ang University\\
Seoul, Republic of Korea \\
\texttt{\{chlwnsgur129, heiscold, seoljh0722, bgnkim\}@cau.ac.kr} \\
}

\begin{document}
\maketitle
\begin{abstract}
We introduce VoiceBBQ\footnote{ \href{https://huggingface.co/datasets/elu-lab/VoiceBBQ}{https://huggingface.co/datasets/bgnkim/VoiceBBQ}}, a spoken extension of the BBQ (Bias Benchmark for Question answering) - a dataset that measures social bias by presenting ambiguous or disambiguated contexts followed by questions that may elicit stereotypical responses. Due to the nature of speech modality, social bias in Spoken Language Models (SLMs) can emerge from two distinct sources: 1) content aspect and 2) acoustic aspect. The dataset converts every BBQ context into controlled voice conditions, enabling per-axis accuracy, bias, and consistency scores that remain comparable to the original text benchmark. Using VoiceBBQ, we evaluate two SLMs—LLaMA-Omni and Qwen2-Audio—and observe architectural contrasts: LLaMA-Omni retains strong acoustic sensitivity, amplifying gender and accent bias, whereas Qwen2-Audio substantially dampens these cues while preserving content fidelity. VoiceBBQ thus provides a compact, drop-in testbed for jointly diagnosing content and acoustic bias across spoken language models.
\end{abstract}

\newcommand{\starx}{\textsuperscript{*}}
\newcommand{\starxx}{\textsuperscript{**}}
\newcommand{\starxxx}{\textsuperscript{***}}

\section{Introduction} 
As the societal influence of AI continues to expand, concerns about social bias in AI systems are growing. Diverse research efforts to detect such bias related to gender or accent have been actively conducted in NLP and CV fields \citep{shrawgi2024uncovering,itzhak2024instructed,VLBiasBench,VLStereoSet,unified,wan-etal-2023-kelly}. But, research on social bias in Spoken Language Models (SLMs) remains relatively limited \citep{lin2024listenspeakfairlystudy, lin2024spoken}, though SLMs have seen a surge in usage recently. As speech modality is widely adopted for real-time interaction, biased responses of SLMs may cause immediate social impact \citep{porcheron2018voice,easwara2015privacy}. Therefore, understanding and mitigating bias in SLMs is crucial for ensuring fair and equitable AI-human interactions.

Due to the nature of speech modality, social bias in SLMs can emerge from two distinct sources: 1) the \textit{content} of utterances and 2) the \textit{acoustic characteristics} of speakers. Nevertheless, most research to date has focused on content-based evaluation \citep{lin2024spoken, lin2024listenspeakfairlystudy}; so, there are not enough reports about whether acoustic characteristics affect social bias. To distinguish the effect of acoustic characteristics from the effect of contents, we need a systematic approach that clearly separates the two sources during the bias assessment.
Therefore, this study aims to conduct a systematic analysis by introducing a benchmark for evaluating both the content and the acoustic aspects of social bias.

Thus, we propose extending a widely used, textual bias benchmark to speech modality. Specifically, we synthesized speech using the Bias Benchmark for Question answering (BBQ; \citet{parrish-etal-2022-bbq}), which evaluates bias by providing contexts about individuals and asking questions where stereotypical assumptions might influence answers. Our experimental design enables controlled quantification of how two primary sources of bias in SLMs, content and acoustic characteristics, influence model behavior by separately evaluating each aspect. For content-related bias, we straightforwardly expanded the method used in BBQ: SLMs answered a text question based on spoken context. For acoustic-related bias, we compared differences in response under four conditions: SLMs received contexts with different gender (male or female) and accent (American or British). 
As a result, our experimental design enables controlled quantification of how two primary sources of bias in SLMs, content and acoustic characteristics, influence model behavior.
Also, by evaluating two SLM architectures, we attempt to draw a hint at how architecture influences bias.

\section{Related Work}
Researchers have examined whether speech processing models have social bias. Early works investigated whether task-specific models, such as speech recognition, suffer from social bias with the acoustic details of speech \cite{Koenecke20ASR, CostaJussa2020, Feng2024ASR, FairSSD, Harris2024ASR}. For example, \citet{Koenecke20ASR} noted that speech recognition systems produced biased results for specific races. Also, \citet{FairSSD} reported that speech synthesis systems generated different outputs for different genders or ages. Furthermore, there is research showing that pretrained speech processing models exhibit human-like biases when performing downstream tasks such as speech emotion recognition \citep{lin2025emo}. While these studies revealed biases in task-specific speech systems such as speech recognition and speech synthesis, they do not evaluate the bias patterns for spoken language models, which serves end-to-end high-level reasoning.

So, researchers recently began to analyze social biases in SLMs \citep{lin2024listenspeakfairlystudy, lin2024spoken}. They especially focused on how contents of a speech affects social biases. For example, \citet{lin2024listenspeakfairlystudy} identified social bias using speech contents when performing tasks as translation, cross-reference resolution, and question-answering. Similarily, \citet{lin2024spoken} conducted an experiment to examine how the content derives bias during a text continuation task. Despite the success of identifying content-related bias in SLMs, they did not evaluate bias affected by acoustic features on speech separately. Their bias evaluation paid less attention to acoustic differences, although these benchmarks has contributed with varying acoustic scenarios, they did not systematically separate how speaker-specific features (e.g., gender and accent) affect bias, which is essential for analyzing acoustic aspect in SLMs.

We believe that distinguishing acoustic property from contents is required in bias evaluation, as SLMs could be affected by both aspects. As we discussed above, acoustic property is essential; Early studies on speech processing studies pointed out acoustic features can affect how the model recognizes input signal. Also, speech content is important; Recent SLM studies mentioned that content can affect social bias in SLMs. However, yet little is known about how these two aspects lead to social bias in SLMs. To achieve a deeper understanding, we need a controlled experiment that could separate the effect of content from that of acoustic properties in evaluating social bias.

\section{VoiceBBQ Benchmark}
We construct a speech variant of the BBQ dataset \citep{parrish-etal-2022-bbq}. By converting the context paragraphs into spoken utterances, we collected 58,492 examples for evaluating auditory social bias. Each instance of BBQ dataset consists of three parts: context, question, and three answer candidates. The original dataset is designed to evaluate social bias across 11 sensitive categories, including gender, race, or socioeconomic status. Though it is possible to convert all parts to speech, we chose the context only. This is because we assumed that longer speech may induce more evident bias if acoustic features actually affect social bias.

\subsection{Speech Synthesis}
To provide a bias benchmark for examining the effect of acoustic features, we synthesized 16 different speeches for each context. We considered two prominent features: gender (male or female) and English accents (American or British). For each combination of gender and accents, we used four different voices because we want to observe average tendencies, not the effect of a specific speaker.

For speech synthesis, we used \texttt{Kokoro-TTS}\footnote{\url{https://huggingface.co/hexgrad/Kokoro-82M}}. We used this model for two reasons. First, the model provides multiple speakers for multiple accents. Second, the model provides a realistic speech based on the StyleTTS2 architecture \citep{li2023styletts}. As \texttt{Kokoro-TTS} does not support long paragraph input, we concatenated sentence-level synthesized results to form the context speech. Appendix \ref{appendix:datasetdetails} further elaborates on the detailed procedure.

As a result, we obtained 935,872 context speech. The average length of speech context was 13.2 seconds, ranging from 2.9 to 40.0 seconds. We further examined whether our context speech mirrors the target acoustic detail; the dataset successfully distinguished target genders and accents. Regarding gender, 96.5\% of context speech showed appropriate acoustic properties when we tested them with a gender classifier \cite{burkhardt2023speechbased}. Similarly, regarding accents, 94.8\% of context speech showed appropriate properties when we tested them with an accent classifier \cite{zuluaga2023commonaccent}.  The details of implementation are in Appendix \ref{appendix:datasetdetails}.

\subsection{Evaluation Metric}
To measure the bias of the SLM, we let SLMs to generate raw responses to the given BBQ item. For each BBQ item, we input speech \textit{context}, \textit{question} and \textit{three answer choices}, and asked them to generate responses. After the generation, we normalized the response and identify the selected option using regular expressions and sequence matching.

The subsequent evaluation followed the original BBQ benchmark procedure \citep{parrish-etal-2022-bbq}. Here, all items were categorized into either the \textit{ambiguous} set or the \textit{disambiguated} set based on predefined criteria. In the ambiguous set, the context lacks sufficient information to determine the right answer; thus, SLMs should respond "UNKNOWN". To evaluate the bias in ambiguous set, BBQ computes \textit{bias score} only when the model chooses answer other than "UNKNOWN." 
In the disambiguated set, the context contains enough information to determine the correct answer.
Different from ambiguous set, BBQ computes \textit{bias score} of disambiguous set by calculating whether SLMs prefer biased option when they respond a non-UNKNOWN answer.

Thus, the focus of BBQ evaluation is not on accuracy; instead, whether the response reflects stereotypical bias is essential. For instance, when a bias score close to zero, it indicates the model has no considerable bias. And, the sign of the bias indicates how much they prefer the biased option. The formula for bias scores are in Appendix \ref{evalprotocol}.

\begin{table*}[!ht]
    \centering
    \small
\begin{tabular}{@{}lr@{\;}r|r@{\;}r|r@{\;}r|r@{\;}r|r@{}l@{\;}r@{}l|r@{}l@{\;}r@{}l@{}}
\toprule
  & \multicolumn{2}{c|}{LLaMA3.1}
  & \multicolumn{2}{c|}{Qwen1}
  & \multicolumn{2}{c|}{LLaMA Omni}
  & \multicolumn{2}{c|}{Qwen2 Audio} & \multicolumn{4}{c|}{LLaMA Omni} 
    & \multicolumn{4}{c@{}}{Qwen2 Audio} \\
\cmidrule(lr){2-3}\cmidrule(lr){4-5}\cmidrule(lr){6-7}\cmidrule(lr){8-9}\cmidrule(lr){10-13}\cmidrule(lr){14-17}
Category & AMB & DIS & AMB & DIS & AMB & DIS & AMB & DIS & \multicolumn{2}{c}{Gender} & \multicolumn{2}{c|}{Accent} & \multicolumn{2}{c}{Gender} & \multicolumn{2}{c@{}}{Accent} \\
\midrule
Age                  &  -0.24 &  -0.33 &  -0.27 &  -0.31 & -0.07 & -0.19 & -0.13 & -0.26 &  68 & {}        &  68 & {}       
    &  35 & {}        &  40 & {} \\

Disability status    &  -0.21 &  -0.32 &  -0.25 &  -0.34 & -0.08 & -0.22 & -0.17 & -0.29 &  26 & {}        &  31 & {}       
    &   8 & {}        &   9 & {} \\

Gender Identity      &  -0.26 &  -0.31 &  -0.31 &  -0.33 & -0.06 & -0.08 & -0.24 & -0.28 & 122 & \starxxx  & 158 & \starx       
    &  70 & {}        &  72 & {} \\

Nationality          & -0.20 & -0.30 & -0.24 & -0.34 & -0.07 & -0.25 & -0.17 & -0.30 &  49 & {}        &  56 & {}       
    &  17 & {}        &  15 & {} \\

Physical Appear.  &  -0.18 &  -0.24 &  -0.15 &  -0.21 & -0.03 & -0.13 & -0.07 & -0.21 &  22 & {}        &  28 & {}       
    &  10 & {}        &  12 & {}  \\

Race/Ethnicity       & -0.20 & -0.29 & -0.25 & -0.28 & -0.09 & -0.28 & -0.20 & -0.29 &  88 & \starx       & 102 & {}       
    &  72 & {}        &  99 & {} \\

Race x SES          & -0.17 & -0.31 & -0.21 & -0.36 & -0.03 & -0.31 & -0.14 & -0.35 &   1 & {}        &   1 & {}       
    &  12 & {}        &  17 & {} \\

Race x Gender       & -0.25 & -0.35 & -0.30 & -0.33 & -0.11 & -0.31 & -0.21 & -0.31 & 233 & {}        & 248 & {}       
    & 139 & {}        & 185 & {}  \\

Religion             & -0.20 & -0.28 & -0.19 & -0.28 & -0.03 & -0.11 & -0.09 & -0.24  &  18 & {}        &  19 & {}       
    &   4 & {}        &   2 & {} \\

SES                  & -0.26 & -0.33 & -0.28 & -0.32 & -0.08 & -0.12 & -0.16 & -0.28 & 200 & \starxxx  & 216 & \starxxx       
    &  47 & {}        &  41 & {} \\

Sexual orient.   & -0.20 & -0.33 & -0.23 & -0.32 & -0.03 & -0.18 & -0.09 & -0.28 &  10 & {}        &   9 & {}       
    &   4 & {}        &   2 & {} \\

\bottomrule
\multicolumn{17}{r}{\starx $p<0.05$, \starxx $p<0.01$, \starxxx $p<0.001$}
\end{tabular}
\caption{Bias scores for two conditions in BBQ, and the result of McNemar test. Appendix \ref{tab:detail} shows detailed results. The 'Gender' and 'Accent' columns show McNemar's chi-square statistics testing whether model responses significantly differ when the same content is spoken by male vs. female voices (Gender) or American vs. British accents (Accent). Higher values indicate greater response variability due to acoustic features.}
\label{tab:transposed_bias_results}
\end{table*}

\section{Experiment}
To systematically evaluate social bias in SLMs, we design our analysis around two key dimensions: content aspect and acoustic aspect. In this section, we describe our analysis methods for examining each bias dimension, and then introduce the two SLM architectures selected for comparison.

 \subsection{Analysis Method}
\paragraph{Content‐Aspect Analysis}
From a content aspect, we suspect that the influence of content on bias in SLMs is largely inherited from the underlying backbone LLM. So, we compare the bias patterns of a SLM with its corresponding backbone LLM. 
To examine whether those two models exhibit similar trends of biased behavior across social categories, we additionally compute Pearson correlation between them. Note that to rule out the effect of acoustic aspect, we averaged results of 16 different voices when analyzing content-aspect biases. Through this analysis, we aim to answer the following question. \\

\noindent \textbf{RQ1}: \textit{Do SLMs have content-induced bias?}

\paragraph{Acoustic Aspect Analysis}
From an acoustic aspect, we aim to examine whether SLMs' predictions vary across speaker conditions such as gender and accent, even with the same input content. We hypothesize that the acoustic features generated by the speech encoder may not fully abstract away speaker-specific attributes before being passed to the LLM. So, the encoder may allow residual acoustic information to influence models' predictions. To test this, we compare predictions across gender and accent conditions and apply McNemar’s test \cite{fagerland2013mcnemar}  to assess whether the differences in decision-making are statistically significant. As we want to make a distinction between biased models, we used disambiguated items that allow different response in biased outputs. Through this analysis, we aim to answer following question. \\

\noindent \textbf{RQ2}: \textit{Does speaker gender or accent affect bias?}

\subsection{Selected Models}
We evaluate two SLMs, LLaMA-Omni \citep{llama-omni} and Qwen2-Audio \citep{qwen2-audio}, along with their respective backbone LLMs. LLaMA-Omni is based on LLaMA 3.1 \citep{grattafiori2024llama} and adopts a modular architecture in which input speech is first processed by a frozen Whisper-large-v3 encoder \citep{whisper}, then passed through a simple speech adapter and the LLM. Second, Qwen2-Audio is based on QwenLM \citep{bai2023qwen} and integrates a Whisper-initialized audio encoder directly into the model's training pipeline. It is trained in an end-to-end manner.

The key distinction lies in how each model handles acoustic information. Because the Whisper encoder is frozen and the speech adapter remains lightweight, the acoustic input is less likely to alter the internal reasoning of the LLM in LLaMA-Omni. In contrast, Qwen2-Audio allows acoustic information such as speaker gender or accent to directly affect the model's semantic representations. As the speech encoder is trained jointly with the model, acoustic characteristics may be preserved and propagated through network.

These architectural differences create contrasting conditions for bias analysis. LLaMA-Omni's modularity allows for relatively independent control over acoustic influence, enabling a clearer attribution of any observed bias to the language model itself rather than to variability in the speech input. This makes LLaMA-Omni suitable for examining whether the model's biases arise from textual understanding rather than acoustic features. Conversely, Qwen2-Audio's design makes it more tightly coupled with the acoustic input, allowing speaker-dependent properties such as gender and accent to affect model predictions, even when the spoken content remains unchanged, making it well-suited for analyzing how variation in vocal delivery influences bias.

\section{Result and Discussion}

\subsection{Content Aspect}
We first compare the two SLM architectures in terms of content-induced bias by examining their relationship with their respective backbone LLMs. Table 1 presents the bias scores for both SLMs and their backbone LLMs across all 11 social categories, along with McNemar test results for acoustic analysis. The left portion shows bias scores for ambiguous (AMB) and disambiguated (DIS) conditions, while the right portion shows McNemar's chi-square statistics testing response variability due to gender and accent.

In response to RQ1, our findings reveal that SLMs do reflect certain content-induced biases observed in their backbone LLMs, but the degree of inheritance varies significantly across architectures. 
Examining the bias score patterns, Qwen2-Audio exhibits a strong correlation with its backbone Qwen1, with a Pearson correlation of r = 0.844 in ambiguous contexts and r = 0.848 in disambiguated contexts. This indicates that Qwen2-Audio largely inherits bias patterns from its underlying language model. This finding aligns with \citet{lin2024listenspeakfairlystudy}, which states that speech-integrated fine tuning reduces some stereotypical associations but does not eliminate content-driven bias.

In contrast, LLaMA-Omni shows a less stable pattern: its correlation with LLaMA 3.1 drops from r = 0.620 in ambiguous contexts to r = 0.301 in disambiguated contexts. This weaker relationship parallels \citet{lin2024spoken} finding that instruction-tuning often reshapes or mixes content biases, with most models exhibiting minimal overall bias yet showing slight stereotypical tendencies in their evaluation. Specifically, in 7 out of 11 categories, LLaMA-Omni shows lower bias scores than its backbone, indicating reduced bias, whereas in the remaining categories, bias increases.

Overall, bias scores in the LLaMA family do not follow a unified direction, in contrast to the Qwen family, which shows consistently increasing bias across all categories. We suspect this difference stems from LLaMA-Omni being trained on the separately constructed InstructS2S-200K dataset, potentially altering inherent biases significantly during training for speech interaction and conciseness. Consequently, LLaMA-Omni notably displayed lower bias scores than Qwen2-Audio across most categories, contradicting initial expectations based solely on backbone comparisons. This demonstrates that biases in LLaMA-Omni were reshaped primarily by the characteristics of the new training data.

\begin{table}[t]
  \centering
  \footnotesize
  \begin{tabular}{@{}l@{\;}|
    c@{}c@{\;}c@{}c@{\;}|
    c@{}c@{\;}c@{}c@{}}
  \toprule
  & \multicolumn{4}{c|}{\textbf{LLaMA Omni}} 
  & \multicolumn{4}{c}{\textbf{Qwen2 Audio}} \\ 
     & \multicolumn{2}{c}{Gender} & \multicolumn{2}{c|}{Accent}
     & \multicolumn{2}{c}{Gender} & \multicolumn{2}{c}{Accent} \\
  & \multicolumn{1}{c}{D} & \multicolumn{1}{c}{A}
  & \multicolumn{1}{c}{D} & \multicolumn{1}{c|}{A}
  & \multicolumn{1}{c}{D} & \multicolumn{1}{c}{A}
  & \multicolumn{1}{c}{D} & \multicolumn{1}{c}{A} \\
  \midrule
  Age                    &  0.5 & 0.5 &  1.2 & 0.3  &  0.3 & 0.2 &  0.5 & -0.5 \\
  Disability status     &  0.3 & 0.5 &  0.9 & 1.5  & -0.9 & -0.3 &  0.9 &  0.3 \\
  Gender identity       &  \textbf{4.0} & 3.0 &  1.5 & 1.2  &  0.1 & 0.1 & -0.7 & -0.6 \\
  Nationality            &  2.8 & 1.4 & -0.9 & 0.4  &  0.4 & 0.4 & -1.3 & -1.3 \\
  Physical Appear.   & -0.5 & 0.2 & \textbf{-5.2} & -1.2  & -0.7 & -0.2 & -2.7 & -1.2 \\
  Race/Ethnicity         &  2.1 & 0.9 &  1.2 & 0.7  & -0.5 & -0.4 &  0.8 & -0.1 \\
  Race x SES           &  \textbf{7.3} & \textbf{4.9} &  1.4 & \textbf{4.9}  & -0.5 & 0.0 & -0.7 &  0.1 \\
  Race x Gender        &  1.1 & 0.7 & -0.2 & 0.1  & -0.1 & 0.0 & -0.5 & -0.2 \\
  Religion               &  \textbf{4.9} & 1.3 &  3.1 & 1.1  & -0.7 & -0.4 &  0.0 & -0.5 \\
  SES                    &  \textbf{5.3} & 3.5 & -1.1 & -0.8 &  0.0 & -0.0 & -0.4 & -0.3 \\
  Sexual orient.    & -0.9 & 0.1 & -3.4 & -0.1 &  3.3 & 1.1 & -2.9 & -0.9 \\
  \bottomrule
  \end{tabular}
  \caption{Bias-score difference ($\Delta s$ in \%) by \textbf{gender} and \textbf{accent} 
for disambiguated (D) vs.\ ambiguous (A) items. Bold values indicate bias score differences 
exceeding 3\%, suggesting substantial influence of acoustic features on model predictions. 
P.A means Physical\_appearance.}
  \label{tab:bias_diff_full}
\end{table}

\subsection{Acoustic Aspect}
We now examine whether acoustic features influence bias patterns by analyzing response variations across different speaker conditions. Table 2 quantifies how bias scores change across acoustic conditions by computing the difference (Δs) between speaker genders and accents for each bias category. The table shows bias score differences (Δs in \%) for disambiguated (D) vs. ambiguous (A) items, with bold values indicating differences exceeding 3\%. The analysis is organized by acoustic dimension: gender effects and accent effects.

In response to RQ2, our findings reveal that the influence of speaker-specific acoustic features—such as gender and accent—varies significantly across SLM architectures. Qwen2-Audio remained stable, exhibiting near-zero bias score differences across gender and accent conditions. In contrast, LLaMA-Omni exhibited significant differences in responses across speaker characteristics.

For gender conditions, significant differences were observed in LLaMA-Omni across multiple categories. In the Gender Identity category (p < 0.001), Race/Ethnicity (p < 0.05), and SES (p < 0.001) categories showed statistically significant variations based on speaker gender, as indicated by McNemar's test results in Table 1.

For accent conditions, significant effects were also found in LLaMA-Omni, particularly in Gender Identity and SES categories, though the statistical significance patterns differ from gender effects.

These outcomes appear to stem from architectural differences between the models. LLaMA-Omni uses a frozen Whisper encoder, whose output is passed through a simple speech adapter to the LLM. As a result, speaker characteristics are transmitted without substantial transformation. In contrast, Qwen2-Audio appears to rely on a Whisper-initialized encoder whose internal representations are shaped in a manner that reduces sensitivity to acoustic variations. This aligns with prior research showing that Whisper produces different responses depending on gender and accent, and such structural modification with diverse speakers can reduce such discrepancies \citep{explore-gender-whisper, Harris2024ASR}.

Consequently, while Qwen2-Audio reduces the impact of acoustic attributes, LLaMA-Omni preserves acoustic features. This allows residual speaker-dependent information to affect models' predictions, leading to observable variation across demographic conditions. These findings complement our content aspect analysis, demonstrating that architectural choices influence both content inheritance and acoustic sensitivity in distinct ways.

\section{Conclusion}
This study introduces Voice BBQ, an extension of the BBQ benchmark for evaluating social bias in SLMs. We analyzed two aspects: content aspect and acoustic aspect.
In content-aspect analysis, we found that Qwen model family transfers bias from backbone to SLMs, while LLaMA family shows a weaker relationship. Notably, LLaMA-Omni, trained on a separate dataset, has lower bias scores.
In acoustic aspect bias analysis, only LLaMA-Omni exhibited significant variations based on speaker characteristics, as it keeps the Whisper encoder frozen. In contrast, Qwen2-Audio's pooling structure dilutes speaker information.

\section*{Limitations}

In this work, we investigated effect of content and acoustics in social bias of SLMs. However, our experiment has three limitations.

First, we were unable to conduct a broad analysis across a wide range of models. Since our experiments were based on open-source SLMs, we had to exclude models that required implementing it from scratch, or those whose available code did not support the desired input-output modalities or failed to run inference in practice. As a result, we employed only two models for our analysis. Further investigation is needed to generalize our findings to other model architectures.

Second, our study primarily focused on diagnosing the content and acoustic biases present in SLMs, without proposing concrete methods for mitigating these biases. As the biases are present in SLMs, we need to reduce such bias to make the model less socially harmful. Therefore, we plan to design and evaluate a SLM architecture that actively mitigates the content and acoustic biases we have identified.

Third, our analysis lacks sufficient exploration of the sociocultural mechanisms underlying the observed acoustic bias patterns. While we identify statistical associations between speaker characteristics (gender, accent) and bias variations, we do not identify a possible theoretical/empirical cause for why these patterns emerge. For instance, why male voices trigger stronger gender identity biases or why certain accent-bias combinations appear remains largely unexplored from sociological and sociolinguistic perspectives.

\section*{Acknowledgments}
We used Grammarly and GPT-4o for polishing our writing. This work was supported by the Institute of Information \& Communications Technology Planning \& Evaluation (IITP) grant funded by the Korea government (MSIT) [RS-2021-II211341, Artificial Intelligence Graduate School Program (Chung-Ang University)].

\bibliography{latex/custom}

\appendix

\section{Details of Dataset}
\label{appendix:datasetdetails}

\subsection{Speech Data Generation}

To construct the spoken version of the BBQ dataset, we synthesized all context sentences using a single TTS model: Kokoro-TTS. This model is based on StyleTTS2 and supports multispeaker synthesis. It can provide speaker voices with varying gender and timbre. For this study, we focused on generating English speech in two regional accents: British (GB) and American (US). Within Kokoro-TTS, we selected predefined speaker voices representing each combination of gender and accent, resulting in a total of 16 unique speakers.

Specifically, the speakers used were as follows:
\begin{itemize}
  \item American Male: \texttt{am\_puck}, \texttt{am\_eric}, \texttt{am\_liam}, \texttt{am\_adam}
  \item American Female: \texttt{af\_heart}, \texttt{af\_sarah}, \texttt{af\_nova}, \texttt{af\_alloy}
  \item British Male: \texttt{bm\_george}, \texttt{bm\_fable}, \texttt{bm\_lewis}, \texttt{bm\_daniel}
  \item British Female: \texttt{bf\_emma}, \texttt{bf\_isabella}, \texttt{bf\_alice}, \texttt{bf\_lily}
\end{itemize}

Each speaker name encodes accent (nationality), gender, and speaker identity, enabling automatic mapping to the appropriate synthesis configuration (e.g., timbre, pitch, speaking style).

For each item in the BBQ dataset, only the context portion was converted into speech using inference (no fine-tuning). When the context contained multiple sentences, we first segmented it using the \texttt{nltk.sent\_tokenize()} function. Each sentence was individually synthesized via Kokoro-TTS and later concatenated using \texttt{numpy.concatenate()} to form a continuous waveform, preserving temporal coherence and natural prosody.

The resulting audio files were saved in 24kHz WAV format. Files were named automatically using metadata such as the question ID and speaker identity, and organized into subfolders by BBQ category. To ensure robustness, the synthesis script was implemented to skip any samples for which the target file already existed, preventing redundant generation.

This process was repeated across all 16 speakers for each of the 58,492 items in the BBQ dataset, yielding a total of 935,872 audio samples with systematically controlled acoustic variation.

The speech synthesis was performed under Python 3.9.21 on a single NVIDIA A6000 ADA GPU. The Kokoro-TTS system was installed by following the official repository\footnote{\url{https://github.com/hexgrad/kokoro}}. The core libraries used for synthesis were: \texttt{torch 2.2.1+cu118}, \texttt{kokoro 0.7.16}, \texttt{pandas 2.2.3}, \texttt{numpy 1.26.4}, \texttt{soundfile 0.13.1}, and \texttt{nltk 3.8.1}.

\subsection{Raw Response Generation Method}
\label{ans_gen_method}

To measure bias in model outputs, we first obtained free-form responses by feeding each synthesized speech file into the evaluation pipeline. Following this, we used a prompt-based mapping method to convert the open-ended responses into structured answer choices, allowing for consistent comparison across conditions.

Specifically, we employed the following prompt template to map each generated response to one of the three multiple-choice options (A, B, or C):

\begin{quote}
You are a classification assistant.\\
Given the context, question, and the model's free-form answer,\\
decide which answer choice (A/B/C) the response most closely matches.\\

\textbf{[Context]}\\
\texttt{\{context\}}\\

\textbf{[Question]}\\
\texttt{\{question\}}\\

\textbf{[Answer Choices]}\\
(A) \texttt{\{ans0\}}\\
(B) \texttt{\{ans1\}}\\
(C) \texttt{\{ans2\}}\\

\textbf{[Model's Answer]}\\
"\texttt{\{llm\_answered\}}"\\

Reply with one letter only: A, B, or C.
\end{quote}


\subsection{BBQ Bias Evaluation Protocol}
\label{evalprotocol}

Following the raw response generation process described in Appendix~\ref{ans_gen_method}, each model output was first normalized by removing all characters except lowercase alphabets, digits, and whitespace. The normalized free-form answer was then compared against the predefined multiple-choice options (\texttt{ans0}, \texttt{ans1}, \texttt{ans2}) from the BBQ benchmark using exact string matching. If no exact match was found, we applied Python's \texttt{SequenceMatcher} to compute string similarity between the model's answer and each candidate option, selecting the index with the highest similarity score.

Once each free-form response was mapped to an answer index (\(\hat{y}_i \in \{0,1,2\}\)), we computed the \textit{accuracy} and \textit{bias score} following the official BBQ evaluation protocol. The formulas are given below:

\[
\mathrm{Accuracy}
=\frac{1}{|\mathcal{D}_{\mathrm{dis}}|}\sum_{i\in\mathcal{D}_{\mathrm{dis}}}\mathbf{1}[\hat{y}_i = y_i]
\]

\[
\mathrm{Bias}
=\frac{\sum_{i\in\mathcal{D}_{\mathrm{und}}} b_i\,\mathbf{1}[\hat{y}_i \neq 2]}
       {\sum_{i\in\mathcal{D}_{\mathrm{und}}} \mathbf{1}[\hat{y}_i \neq 2]}
\]

Here, \(\mathcal{D}_{\mathrm{dis}}\) denotes the set of disambiguated items and \(\mathcal{D}_{\mathrm{und}}\) denotes the set of ambiguous items. \(y_i\) is the ground-truth label, \(b_i\) is the bias indicator (i.e., which choice reflects a stereotyped response), and \(\hat{y}_i\) is the model’s predicted choice index. These definitions follow the official BBQ benchmark metrics, allowing direct comparability with prior studies \citep{parrish-etal-2022-bbq}.

\subsection{Data Validation Process}

The speaker metadata validation step was conducted under Python 3.10.16 using a single NVIDIA A6000 ADA GPU. The setup followed the configuration guidelines provided on the Hugging Face model pages\footnote{\url{https://huggingface.co/audeering/wav2vec2-large-robust-24-ft-age-gender}}\footnote{\url{https://huggingface.co/Jzuluaga/accent-id-commonaccent_ecapa}}.

\begin{table}[ht]
\small
\centering
\begin{tabular}{lrrr}
\toprule
Ground-Truth & Total Samples & Accuracy (\%) \\
\midrule
Gender: Female & 467{,}936 & 100.00 \\
Gender: Male  & 467{,}936 & 93.16 \\
Region: GB  & 467{,}936 & 99.49 \\
Region: US  & 467{,}936 & 90.15 \\
\bottomrule
\end{tabular}
\caption{Prediction accuracy by ground-truth category (GB and US denote Great Britain and United States).}
\label{tab:gender_region_accuracy}
\end{table}

We employed two pretrained audio classification models based on different architectures. The first, \texttt{wav2vec2-large-robust-24-ft-age-gender}, is a wav2vec 2.0–based model fine-tuned for age and gender classification after pretraining on large-scale datasets such as VoxCeleb. According to its official documentation, it achieves over 80\% balanced accuracy on gender classification tasks. The second, \texttt{accent-id-commonaccent\_ecapa}, is built upon the ECAPA-TDNN architecture and was trained on the CommonVoice dataset to identify English regional accents, reporting classification accuracy above 90\%.

These models were chosen due to their verified performance on downstream tasks relevant to our study—namely, speaker gender and accent (region) classification—which made them suitable for validating the integrity of the synthesized speech data. Each model was used to infer gender or region from the input audio waveform under \texttt{eval()} mode with batch size 1.

The major libraries used for inference were: \texttt{torch 2.5.1+cu124}, \texttt{transformers 4.51.3}, \texttt{numpy 1.26.4}, and \texttt{pandas 2.2.3}.

\section{Experimental Environment}
In this section provides a concise overview of the hardware configurations, software setups, and library dependencies used in our Qwen2-Audio and LLaMA-Omni experiments.

\subsection{Qwen2-Audio}
All Qwen2-Audio experiments were conducted on a single NVIDIA A6000 GPU under Python 3.9.21. The environment was configured following the model’s Hugging Face page\footnote{\url{https://huggingface.co/Qwen/Qwen2-Audio-7B-Instruct}}. Inference was performed using the Hugging Face \texttt{AutoProcessor} and \texttt{Qwen2AudioForConditionalGeneration}, jointly processing text and audio inputs and generating outputs via: \texttt{model.generate(max\_length=1024)} To ensure comparability across runs, the maximum token length was fixed at 1024, and all experiments were executed with batch size 1 in evaluation mode \texttt{eval()}. Major dependencies included \texttt{torch 2.2.1+cu118}, \texttt{transformers 4.52.0}, \texttt{numpy 2.2.5}, \texttt{pandas 2.2.3}, and \texttt{soundfile 0.13.1}.

\subsection{LLaMA-Omni}
LLaMA-Omni experiments were carried out on a single A6000 GPU under Python 3.10.17. The setup was based on OmniMMI’s OpenOmniNexus framework\footnote{\url{https://github.com/OmniMMI/OpenOmniNexus})} and the official LLaMA-Omni repository\footnote{\url{https://github.com/ictnlp/LLaMA-Omni}}. Model and tokenizer were loaded using : \texttt{load\_pretrained\_model(model\_path, None, s2s=False)} Inference was performed with batch size 1 in evaluation mode, using \texttt{do\_sample=False}, \texttt{num\_beams=1}, \texttt{top\_p=None}, and \texttt{max\_new\_tokens=1024}. For fairness, the maximum token count was fixed at 1024. Major dependencies included \texttt{torch 2.5.0+cu118}, \texttt{transformers 4.44.0}, \texttt{flash-attn 2.6.3}, \texttt{fairseq 0.12.2}, \texttt{deepspeed 0.14.5}, and \texttt{numpy 1.26.4}.

\begin{table}[!t]
\small
\centering
\begin{tabular}{lrrr}
\toprule
Comparison model & Context & r & p-value \\
\midrule
LLaMA-based & Ambiguous & 0.620 & 0.042 \\
LLaMA-based & Disambiguated & 0.301 & 0.369 \\
Qwen-based & Ambiguous & 0.844 & p < 0.001 \\
Qwen-based & Disambiguated & 0.848 & p < 0.001 \\
\bottomrule
\end{tabular}
\caption{Pearson correlation analysis for LLaMA family and Qwen family for ambiguous and disambiguated context.}
\label{tab:pearson_corr}
\end{table}

\subsection{Backbone LLM Model Setting}

The BBQ benchmark evaluation of the backbone LLMs was performed under Python 3.10.16 using a single NVIDIA A6000 ADA GPU. The experimental setup followed the official Hugging Face configurations for each model: \texttt{Qwen/Qwen-7B}\footnote{\url{https://huggingface.co/Qwen/Qwen-7B}} and \texttt{meta-llama/Llama-3.1-8B-Instruct}\footnote{\url{https://huggingface.co/meta-llama/Llama-3.1-8B-Instruct}}.

Inference was conducted using \texttt{AutoTokenizer} and \texttt{AutoModelForCausalLM}. Text inputs were passed to the models and generation was performed using model.generate(max\_new\_tokens=1024). To ensure consistency across runs, the maximum number of tokens was fixed at 1024. All inference runs were executed in evaluation mode (\texttt{eval()}) with batch size set to 1.

The main libraries used for this process included: \texttt{torch 2.5.1+cu124}, \texttt{transformers 4.51.3}, \texttt{numpy 1.26.4}, and \texttt{pandas 2.2.3}.

\begin{table*}[!t]
    \centering
    \small
    \resizebox{\textwidth}{!}{
\begin{tabular}{@{}l   rrrr|rrrr|rrrr|rrrr@{}}
\toprule
  & \multicolumn{4}{c|}{LLaMA3.1}
  & \multicolumn{4}{c|}{Qwen1}
  & \multicolumn{4}{c|}{LLaMA Omni}
  & \multicolumn{4}{c}{Qwen2 Audio} \\
\cmidrule(lr){2-5}\cmidrule(lr){6-9}\cmidrule(lr){10-13}\cmidrule(lr){14-17}
Category & \textrm{acc}\_A & \textrm{acc}\_D & s\_A & s\_D & \textrm{acc}\_A & \textrm{acc}\_D & s\_A & s\_D & \textrm{acc}\_A & \textrm{acc}\_D & s\_A & s\_D & \textrm{acc}\_A & \textrm{acc}\_D & s\_A & s\_D \\
\midrule
Age                  & 0.258 & 0.793 &  -0.242 &  -0.326 & 0.131 & 0.852 &   -0.266 &  -0.306 & 0.655 & 0.536 &  -0.065 & -0.187 & 0.494 & 0.803 &  -0.132 & -0.260 \\

Disability status    & 0.337 & 0.867 &  -0.21 &  -0.317 & 0.275 & 0.942 &   -0.247 &  -0.341 & 0.646 & 0.644 &  -0.076 & -0.216 & 0.423 & 0.899 &  -0.169 & -0.293 \\

Gender Identity      & 0.17 & 0.634 &  -0.255 &  -0.308 & 0.081 & 0.823  &   -0.307 &  -0.334 & 0.251 & 0.558 &  -0.061 & -0.081 & 0.133 & 0.824 &  -0.241 & -0.278 \\

Nationality          & 0.352 & 0.909 & -0.197 & -0.303 & 0.286 & 0.903 &  -0.244 & -0.342 & 0.735 & 0.554 &  -0.065 & -0.246 & 0.428 & 0.758 &  -0.168 & -0.294 \\

Physical Appearance  & 0.271 & 0.738 &  -0.177 &  -0.243 & 0.299 & 0.782 &   -0.15 &  -0.214 & 0.744 & 0.545 &  -0.033 & -0.128 & 0.653 & 0.718 &  -0.074 & -0.213 \\

Race/Ethnicity       & 0.322 & 0.722 & -0.196 & -0.289 & 0.096 & 0.793 &  -0.252 & -0.279 & 0.684 & 0.534 &  -0.089 & -0.281 & 0.321 & 0.812 &  -0.197 & -0.291 \\

Race x SES          & 0.448 & 0.828 & -0.17 & -0.309 & 0.434 & 0.993 &  -0.206 & -0.363 & 0.918 & 0.622 &  -0.025 & -0.305 & 0.619 & 0.946 &  -0.135 & -0.354 \\

Race x gender       & 0.271 & 0.743 & -0.254 & -0.348 & 0.1 & 0.844 &  -0.297 & -0.33 & 0.642 & 0.655 &  -0.110 & -0.306 & 0.331 & 0.874 &  -0.206 & -0.308 \\

Religion             & 0.287 & 0.837 & -0.202 & -0.283 & 0.314 & 0.860 &  -0.190 & -0.277 & 0.776 & 0.553 &  -0.025 & -0.112 & 0.636 & 0.927 &  -0.087 & -0.240 \\

SES                  & 0.207 & 0.871 & -0.263 & -0.331 & 0.102 & 0.952 &  -0.283 & -0.315 & 0.368 & 0.554 &  -0.075 & -0.119 & 0.438 & 0.912 &  -0.155 & -0.276 \\

Sexual orientation   & 0.415 & 0.826 & -0.195 & -0.333 & 0.288 & 0.832 &  -0.225 & -0.315 & 0.839 & 0.584 &  -0.029 & -0.181 & 0.676 & 0.854 &  -0.091 & -0.281 \\

\bottomrule
\end{tabular}
}
\caption{Accuracy and Bias for disambiguated ($\textrm{acc}_D$, $\textrm{s}_D$) vs. ambiguous ($\textrm{acc}_A$, $\textrm{s}_A$) items. 
         $\textrm{a}_D$, $\textrm{a}_A$ denote accuracy, and $\textrm{s}_D$, $\textrm{s}_A$ denote bias score.}
\label{tab:accuracy_bias_results1}
\end{table*}

\begin{table*}[!t]
  \centering
  \small
  \resizebox{\textwidth}{!}{%
  \begin{tabular}{
    @{}l@{\;\;}|
    *{8}{c@{\;\;}}|
    *{8}{c@{\;\;}}@{}
    }
  
  \toprule
  & \multicolumn{8}{c|}{\textbf{LLaMA-Omni}} 
  & \multicolumn{8}{c}{\textbf{Qwen2-Audio}} \\ 
     & \multicolumn{4}{c}{Female} & \multicolumn{4}{c|}{Male}
     & \multicolumn{4}{c}{Female} & \multicolumn{4}{c}{Male} \\
  & \multicolumn{1}{c}{acc$_D$} & \multicolumn{1}{c}{acc$_A$} &\multicolumn{1}{c}{s$_D$} & \multicolumn{1}{c}{s$_A$}
  & \multicolumn{1}{c}{acc$_D$} & \multicolumn{1}{c}{acc$_A$} &\multicolumn{1}{c}{s$_D$} & \multicolumn{1}{c}{s$_A$}
  & \multicolumn{1}{c}{acc$_D$} & \multicolumn{1}{c}{acc$_A$} &\multicolumn{1}{c}{s$_D$} & \multicolumn{1}{c}{s$_A$}
  & \multicolumn{1}{c}{acc$_D$} & \multicolumn{1}{c}{acc$_A$} &\multicolumn{1}{c}{s$_D$} & \multicolumn{1}{c}{s$_A$}  \\
    \midrule
    Age                    & 0.537 & 0.647 & -0.190 & -0.067 & 0.534 & 0.663 & -0.184 & -0.062 & 0.809 & 0.493 & -0.262 & -0.133 & 0.797 & 0.495 & -0.258 & -0.130 \\
    Disability\_status     & 0.647 & 0.636 & -0.217 & -0.079 & 0.641 & 0.657 & -0.214 & -0.074 & 0.896 & 0.418 & -0.289 & -0.168 & 0.901 & 0.427 & -0.298 & -0.171 \\
    Gender\_identity       & 0.564 & 0.254 & -0.101 & -0.075 & 0.552 & 0.249 & -0.061 & -0.046 & 0.823 & 0.133 & -0.278 & -0.241 & 0.824 & 0.133 & -0.277 & -0.240 \\
    Nationality            & 0.553 & 0.721 & -0.259 & -0.072 & 0.555 & 0.749 & -0.231 & -0.058 & 0.757 & 0.425 & -0.296 & -0.170 & 0.759 & 0.431 & -0.292 & -0.166 \\
    Physical\_appearance   & 0.556 & 0.733 & -0.126 & -0.034 & 0.534 & 0.756 & -0.130 & -0.032 & 0.724 & 0.653 & -0.209 & -0.073 & 0.711 & 0.653 & -0.216 & -0.075 \\
    Race\_ethnicity        & 0.539 & 0.680 & -0.291 & -0.093 & 0.529 & 0.688 & -0.271 & -0.084 & 0.810 & 0.323 & -0.288 & -0.195 & 0.814 & 0.319 & -0.293 & -0.199 \\
    Race\_x\_SES           & 0.632 & 0.839 & -0.308 & -0.050 & 0.376 & 0.997 & -0.235 & -0.001 & 0.945 & 0.616 & -0.352 & -0.135 & 0.946 & 0.622 & -0.357 & -0.135 \\
    Race\_x\_gender        & 0.652 & 0.636 & -0.311 & -0.113 & 0.657 & 0.647 & -0.300 & -0.106 & 0.875 & 0.329 & -0.307 & -0.206 & 0.873 & 0.333 & -0.309 & -0.206 \\
    Religion               & 0.557 & 0.769 & -0.137 & -0.032 & 0.549 & 0.783 & -0.087 & -0.019 & 0.926 & 0.639 & -0.237 & -0.086 & 0.928 & 0.633 & -0.243 & -0.089 \\
    SES                    & 0.564 & 0.360 & -0.145 & -0.093 & 0.544 & 0.376 & -0.092 & -0.057 & 0.909 & 0.439 & -0.276 & -0.155 & 0.915 & 0.437 & -0.276 & -0.155 \\
    Sexual\_orientation     & 0.585 & 0.833 & -0.176 & -0.029 & 0.584 & 0.845 & -0.185 & -0.029 & 0.858 & 0.676 & -0.297 & -0.096 & 0.851 & 0.676 & -0.264 & -0.085 \\

\bottomrule

  \end{tabular}
}
  \caption{Accuracy and Bias-score by speaker \textbf{Gender}
           for disambiguated ($\textrm{acc}_D$, $s_D$) vs. ambiguous ($\textrm{acc}_A$, $\textrm{acc}_A$) items. $\textrm{acc}_D$,  $\textrm{acc}_A$ denote accuracy, and $s_D$, $s_A$ denote bias score.}
  \label{tab:bias_diff_full_gender}
\end{table*}

\begin{table*}[!t]
  \centering
  \small
  \resizebox{\textwidth}{!}{%
  \begin{tabular}{
    @{}l@{\;\;}|
    *{8}{c@{\;\;}}|
    *{8}{c@{\;\;}}@{}
    }
  
  \toprule
  & \multicolumn{8}{c|}{\textbf{LLaMA-Omni}} 
  & \multicolumn{8}{c}{\textbf{Qwen2-Audio}} \\ 
     & \multicolumn{4}{c}{US} & \multicolumn{4}{c|}{UK}
     & \multicolumn{4}{c}{US} & \multicolumn{4}{c}{UK} \\
  & \multicolumn{1}{c}{acc\_D} & \multicolumn{1}{c}{acc\_A} &\multicolumn{1}{c}{s\_D} & \multicolumn{1}{c}{s\_A}
  & \multicolumn{1}{c}{acc\_D} & \multicolumn{1}{c}{acc\_A} &\multicolumn{1}{c}{s\_D} & \multicolumn{1}{c}{s\_A}
  & \multicolumn{1}{c}{acc\_D} & \multicolumn{1}{c}{acc\_A} &\multicolumn{1}{c}{s\_D} & \multicolumn{1}{c}{s\_A}
  & \multicolumn{1}{c}{acc\_D} & \multicolumn{1}{c}{acc\_A} &\multicolumn{1}{c}{s\_D} & \multicolumn{1}{c}{s\_A}  \\
    \midrule
    Age                   & 0.539 & 0.652 & -0.181 & -0.062 & 0.532 & 0.657 & -0.193 & -0.066 & 0.807 & 0.479 & -0.258 & -0.134 & 0.799 & 0.509 & -0.262 & -0.129 \\
    Disability\_status    & 0.635 & 0.673 & -0.211 & -0.069 & 0.653 & 0.620 & -0.220 & -0.084 & 0.894 & 0.420 & -0.289 & -0.168 & 0.903 & 0.426 & -0.298 & -0.171 \\
    Gender\_identity       & 0.556 & 0.254 & -0.073 & -0.055 & 0.561 & 0.249 & -0.088 & -0.066 & 0.826 & 0.133 & -0.281 & -0.244 & 0.822 & 0.134 & -0.274 & -0.238 \\
    Nationality           & 0.553 & 0.747 & -0.250 & -0.063 & 0.556 & 0.722 & -0.242 & -0.067 & 0.751 & 0.419 & -0.301 & -0.175 & 0.765 & 0.437 & -0.287 & -0.162 \\
    Physical\_appearance   & 0.547 & 0.748 & -0.154 & -0.039 & 0.544 & 0.741 & -0.102 & -0.027 & 0.724 & 0.647 & -0.226 & -0.080 & 0.712 & 0.659 & -0.199 & -0.068 \\
    Race/Ethnicity         & 0.534 & 0.691 & -0.275 & -0.085 & 0.535 & 0.678 & -0.287 & -0.093 & 0.812 & 0.311 & -0.287 & -0.198 & 0.813 & 0.331 & -0.294 & -0.197 \\
    Race\_x\_SES            & 0.389 & 0.997 & -0.291 & -0.001 & 0.634 & 0.839 & -0.306 & -0.049 & 0.947 & 0.624 & -0.358 & -0.134 & 0.944 & 0.613 & -0.351 & -0.136 \\
    Race\_x\_gender         & 0.652 & 0.645 & -0.307 & -0.109 & 0.657 & 0.639 & -0.305 & -0.110 & 0.874 & 0.334 & -0.311 & -0.207 & 0.874 & 0.329 & -0.305 & -0.205 \\
    Religion              & 0.560 & 0.795 & -0.097 & -0.020 & 0.546 & 0.757 & -0.127 & -0.031 & 0.926 & 0.625 & -0.240 & -0.090 & 0.929 & 0.647 & -0.240 & -0.085 \\
    SES                    & 0.558 & 0.362 & -0.124 & -0.079 & 0.550 & 0.373 & -0.113 & -0.071 & 0.913 & 0.437 & -0.278 & -0.157 & 0.911 & 0.439 & -0.274 & -0.154 \\
    Sexual\_orientation     & 0.580 & 0.850 & -0.197 & -0.030 & 0.588 & 0.828 & -0.163 & -0.028 & 0.861 & 0.677 & -0.295 & -0.095 & 0.848 & 0.676 & -0.266 & -0.086 \\
    
\bottomrule

  \end{tabular}
}
  \caption{Accuracy and Bias-score by speaker \textbf{Accent}
           for disambiguated ($\textrm{acc}_D$, $s_D$) vs. ambiguous ($\textrm{acc}_A$, $s_A$) items. $\textrm{acc}_D$, $\textrm{acc}_A$ denote accuracy, and $s_D$, $s_A$ denote bias score.}
  \label{tab:bias_diff_full_national}
\end{table*}

\section{Detail Result}
\label{tab:detail}
Tables from \ref{tab:accuracy_bias_results1} to \ref{tab:bias_diff_full_national} shows the detailed result of Bias Socre, and table \ref{tab:pearson_corr} show the detailed result of correlation result and std value.

\begin{table*}[]
\centering
\begin{tabular}{lcccc}
\hline
\textbf{Category} & \textbf{$s_D$ Std} & \textbf{$\textrm{acc}_D$ Std} & \textbf{$\textrm{acc}_A$ Std} & \textbf{$s_A$ Std} \\
\hline
Age                & 0.021 & 0.017 & 0.035 & 0.015 \\
Disability\_status & 0.016 & 0.010 & 0.034 & 0.010 \\
Gender\_identity   & 0.009 & 0.008 & 0.011 & 0.008 \\
Nationality        & 0.016 & 0.014 & 0.025 & 0.014 \\
Physical\_appearance & 0.027 & 0.020 & 0.030 & 0.013 \\
Race\_ethnicity    & 0.009 & 0.009 & 0.022 & 0.010 \\
Race\_x\_SES       & 0.009 & 0.008 & 0.021 & 0.008 \\
Race\_x\_gender    & 0.005 & 0.005 & 0.031 & 0.011 \\
Religion           & 0.014 & 0.008 & 0.029 & 0.009 \\
SES                & 0.009 & 0.008 & 0.019 & 0.008 \\
Sexual\_orientation & 0.039 & 0.021 & 0.031 & 0.018 \\
\hline
\end{tabular}
\caption{Standard Deviation of Qwen Audio, Global}
\end{table*}

\begin{table*}[]
\centering

\begin{tabular}{lcccccccc}
\hline
\textbf{Category} & 
\multicolumn{2}{c}{\textbf{$s_D$ Std}} & 
\multicolumn{2}{c}{\textbf{$\textrm{acc}_D$ Std}} & 
\multicolumn{2}{c}{\textbf{$\textrm{acc}_A$ Std}} & 
\multicolumn{2}{c}{\textbf{$s_A$ Std}} \\
\cline{2-9}
 & female & male & female & male & female & male & female & male \\
\hline
Age                & 0.027 & 0.015 & 0.020 & 0.013 & 0.040 & 0.033 & 0.019 & 0.011 \\
Disability\_status & 0.017 & 0.014 & 0.008 & 0.010 & 0.034 & 0.035 & 0.011 & 0.009 \\
Gender\_identity   & 0.009 & 0.008 & 0.010 & 0.007 & 0.013 & 0.010 & 0.009 & 0.008 \\
Nationality        & 0.016 & 0.017 & 0.013 & 0.015 & 0.024 & 0.028 & 0.015 & 0.014 \\
Physical\_appearance & 0.036 & 0.018 & 0.017 & 0.021 & 0.028 & 0.035 & 0.016 & 0.009 \\
Race\_ethnicity    & 0.009 & 0.009 & 0.011 & 0.006 & 0.025 & 0.020 & 0.010 & 0.010 \\
Race\_x\_SES       & 0.009 & 0.009 & 0.004 & 0.011 & 0.013 & 0.027 & 0.007 & 0.010 \\
Race\_x\_gender    & 0.006 & 0.004 & 0.006 & 0.004 & 0.034 & 0.030 & 0.013 & 0.010 \\
Religion           & 0.013 & 0.014 & 0.008 & 0.008 & 0.027 & 0.032 & 0.007 & 0.010 \\
SES                & 0.011 & 0.007 & 0.007 & 0.008 & 0.019 & 0.021 & 0.007 & 0.008 \\
Sexual\_orientation & 0.045 & 0.026 & 0.021 & 0.022 & 0.026 & 0.036 & 0.021 & 0.013 \\
\hline
\end{tabular}
\caption{Standard Deviation of Qwen Audio, Gender}
\end{table*}

\begin{table*}[]
\centering

\begin{tabular}{lcccccccc}
\hline
\textbf{Category} & 
\multicolumn{2}{c}{\textbf{$s_D$ Std}} & 
\multicolumn{2}{c}{\textbf{$\textrm{acc}_D$ Std}} & 
\multicolumn{2}{c}{\textbf{$\textrm{acc}_A$ Std}} & 
\multicolumn{2}{c}{\textbf{$s_A$ Std}} \\
\cline{2-9}
 & GB & US & GB & US & GB & US & GB & US \\
\hline
Age                & 0.019 & 0.024 & 0.013 & 0.021 & 0.037 & 0.027 & 0.015 & 0.016 \\
Disability\_status & 0.018 & 0.013 & 0.008 & 0.009 & 0.038 & 0.031 & 0.010 & 0.009 \\
Gender\_identity   & 0.006 & 0.010 & 0.009 & 0.008 & 0.010 & 0.013 & 0.007 & 0.009 \\
Nationality        & 0.015 & 0.014 & 0.014 & 0.009 & 0.022 & 0.027 & 0.014 & 0.012 \\
Physical\_appearance & 0.018 & 0.029 & 0.017 & 0.022 & 0.035 & 0.026 & 0.011 & 0.012 \\
Race\_ethnicity    & 0.008 & 0.009 & 0.010 & 0.008 & 0.021 & 0.019 & 0.011 & 0.010 \\
Race\_x\_SES       & 0.009 & 0.009 & 0.006 & 0.010 & 0.020 & 0.022 & 0.009 & 0.008 \\
Race\_x\_gender    & 0.003 & 0.004 & 0.004 & 0.007 & 0.031 & 0.033 & 0.011 & 0.012 \\
Religion           & 0.012 & 0.016 & 0.008 & 0.008 & 0.025 & 0.030 & 0.008 & 0.010 \\
SES                & 0.010 & 0.008 & 0.006 & 0.010 & 0.020 & 0.021 & 0.005 & 0.009 \\
Sexual\_orientation & 0.028 & 0.045 & 0.025 & 0.015 & 0.035 & 0.028 & 0.014 & 0.020 \\
\hline
\end{tabular}
\caption{Standard Deviation of Qwen Audio, Accent}
\end{table*}

\begin{table*}[]
\centering
\begin{tabular}{lcccc}
\hline
\textbf{Category} & \textbf{$s_D$ Std} & \textbf{$\textrm{acc}_D$ Std} & \textbf{$\textrm{acc}_A$ Std} & \textbf{$s_A$ Std} \\
\hline
Age                & 0.022 & 0.009 & 0.015 & 0.009 \\
Disability\_status & 0.037 & 0.022 & 0.045 & 0.014 \\
Gender\_identity   & 0.026 & 0.010 & 0.012 & 0.020 \\
Nationality        & 0.036 & 0.016 & 0.029 & 0.013 \\
Physical\_appearance & 0.051 & 0.022 & 0.015 & 0.013 \\
Race\_ethnicity    & 0.017 & 0.007 & 0.018 & 0.008 \\
Race\_x\_SES       & 0.134 & 0.136 & 0.141 & 0.044 \\
Race\_x\_gender    & 0.011 & 0.005 & 0.009 & 0.006 \\
Religion           & 0.062 & 0.033 & 0.028 & 0.016 \\
SES                & 0.046 & 0.017 & 0.021 & 0.031 \\
Sexual\_orientation & 0.045 & 0.023 & 0.022 & 0.008 \\
\hline
\end{tabular}
\caption{Standard Deviation of LLaMA-Omni, Global}
\end{table*}

\begin{table*}[]
\centering
\begin{tabular}{lcccccccc}
\hline
\textbf{Category} & 
\multicolumn{2}{c}{\textbf{$s_D$ Std}} & 
\multicolumn{2}{c}{\textbf{$\textrm{acc}_D$ Std}} & 
\multicolumn{2}{c}{\textbf{$\textrm{acc}_A$Std}} & 
\multicolumn{2}{c}{\textbf{$s_A$ Std}} \\
\cline{2-9}
 & female & male & female & male & female & male & female & male \\
\hline
Age                & 0.024 & 0.022 & 0.010 & 0.009 & 0.016 & 0.011 & 0.009 & 0.009 \\
Disability\_status & 0.032 & 0.043 & 0.028 & 0.016 & 0.059 & 0.024 & 0.012 & 0.017 \\
Gender\_identity   & 0.015 & 0.018 & 0.009 & 0.008 & 0.010 & 0.014 & 0.012 & 0.013 \\
Nationality        & 0.037 & 0.031 & 0.016 & 0.018 & 0.030 & 0.021 & 0.012 & 0.011 \\
Physical\_appearance & 0.049 & 0.057 & 0.024 & 0.014 & 0.009 & 0.011 & 0.013 & 0.013 \\
Race\_ethnicity    & 0.015 & 0.011 & 0.006 & 0.003 & 0.020 & 0.017 & 0.007 & 0.007 \\
Race\_x\_SES       & 0.104 & 0.160 & 0.128 & 0.098 & 0.169 & 0.001 & 0.053 & 0.000 \\
Race\_x\_gender    & 0.009 & 0.010 & 0.004 & 0.005 & 0.006 & 0.008 & 0.004 & 0.005 \\
Religion           & 0.059 & 0.059 & 0.036 & 0.030 & 0.030 & 0.026 & 0.017 & 0.014 \\
SES                & 0.031 & 0.045 & 0.017 & 0.011 & 0.013 & 0.025 & 0.021 & 0.029 \\
Sexual\_orientation & 0.046 & 0.047 & 0.025 & 0.022 & 0.029 & 0.011 & 0.010 & 0.007 \\
\hline
\end{tabular}
\caption{Standard Deviation of LLaMA Omni, Gender}
\end{table*}

\begin{table*}[]
\centering
\begin{tabular}{lcccccccc}
\hline
\textbf{Category} & 
\multicolumn{2}{c}{\textbf{$s_D$ Std}} & 
\multicolumn{2}{c}{\textbf{$\textrm{acc}_D$ Std}} & 
\multicolumn{2}{c}{\textbf{$\textrm{acc}_A$ Std}} & 
\multicolumn{2}{c}{\textbf{$s_A$ Std}} \\
\cline{2-9}
 & gb & us & gb & us & gb & us & gb & us \\
\hline
Age                & 0.023 & 0.022 & 0.010 & 0.008 & 0.007 & 0.021 & 0.009 & 0.009 \\
Disability\_status & 0.035 & 0.040 & 0.019 & 0.022 & 0.050 & 0.014 & 0.012 & 0.014 \\
Gender\_identity   & 0.025 & 0.027 & 0.007 & 0.012 & 0.010 & 0.013 & 0.019 & 0.020 \\
Nationality        & 0.029 & 0.044 & 0.015 & 0.018 & 0.028 & 0.026 & 0.011 & 0.015 \\
Physical\_appearance & 0.054 & 0.035 & 0.015 & 0.029 & 0.015 & 0.016 & 0.014 & 0.009 \\
Race\_ethnicity    & 0.014 & 0.018 & 0.007 & 0.007 & 0.020 & 0.015 & 0.008 & 0.007 \\
Race\_x\_SES       & 0.107 & 0.160 & 0.177 & 0.034 & 0.169 & 0.001 & 0.053 & 0.000 \\
Race\_x\_gender    & 0.011 & 0.011 & 0.006 & 0.004 & 0.007 & 0.009 & 0.005 & 0.007 \\
Religion           & 0.070 & 0.053 & 0.031 & 0.034 & 0.022 & 0.019 & 0.019 & 0.012 \\
SES                & 0.039 & 0.055 & 0.011 & 0.022 & 0.025 & 0.015 & 0.026 & 0.036 \\
Sexual\_orientation & 0.049 & 0.036 & 0.020 & 0.026 & 0.024 & 0.014 & 0.011 & 0.005 \\
\hline
\end{tabular}
\caption{Standard Deviation of LLaMA Omni, Accent}
\end{table*}

\end{document}